\documentclass{llncs}

\usepackage{epsf}
\usepackage{cite}
\usepackage{graphicx}

\usepackage{epstopdf}
\usepackage{color}
\usepackage[utf8x]{inputenc}
\usepackage{textcomp}
\usepackage[T1]{fontenc}
\usepackage{fancybox}

\begin{document}
\title{Sentiment Analysis in Scholarly Book Reviews}

\author{
Hussam Hamdan*,**,***, Patrice Bellot*,**, Frederic Bechet***}
\institute{*Aix Marseille Université, CNRS, ENSAM, Universit{é} de Toulon, LSIS UMR 7296,13397, Marseille, France {hussam.hamdan,patrice.bellot}@lsis.org \\ *** Aix Marseille Université, CNRS, LIF UMR 7279, Marseille, France {frederic.bechet}@lif.univ-mrs.fr }
\maketitle

\begin{abstract}
So far different studies have tackled the sentiment analysis in several domains such as restaurant and movie reviews. But, this problem has not been studied in scholarly book reviews which is different in terms of review style and size. In this paper, we propose to combine different features in order to be presented to a supervised classifiers which extract the opinion target expressions and detect their polarities in scholarly book reviews. We construct a labeled corpus for training and evaluating our methods in French book reviews. We also evaluate them on English restaurant reviews in order to measure their robustness across the domains and  languages. The evaluation shows that our methods are enough robust for English restaurant reviews and  French  book reviews.

\end{abstract}

\section{Introduction}
Classifying opinion texts at document or sentence levels is not sufficient for applications which need to identify the opinion targets. Even if the document is about one entity, many applications need to determine the opinion about each aspect of the entity. A user may express a positive opinion towards the food in a restaurant, but he may have a negative opinion towards other aspects as the ambiance. Therefore, we need to identify the aspects and determine whether the sentiment is positive, negative or neutral towards each one. This task is called Aspect-Based Sentiment Analysis or Feature-Based opinion mining as called in the early work \cite{hu_mining_2004}.

In this work, we address the problem of sentiment analysis in scholarly book reviews. Our objective is to extract the opinion expressed towards a book in all its reviews. Therefore, given a collection of book reviews, we aim at finding out the aspects of the book and the sentiment expressed towards each aspect. This seems similar to aspect-based sentiment analysis in restaurant reviews where we have a set of aspects such as (food, drinks, service, ambiance, location), each aspect involves different aspect terms or opinion target expressions i.g. Pizza, Burger in food aspect.

While it is not difficult to have a list of aspects in restaurant domain, it is ambiguous what may be the aspects of a book. When one thinks about the aspects of books, he may think about the quality of book, the number of pages, the discussed topics ...etc. But it is still not obvious as in restaurant reviews where all people may consider without doubt the food and drinks as aspects or categories. In fact, one can consider two methods to determine the aspects:
\begin{enumerate}
  \item 
Applying unsupervised method which is capable of extracting the facets or topics such as topic modeling in which we consider each topic related to an aspect. It is not obvious how we can evaluate the quality of this method and how each topic related to an aspect.
  \item 
Asking domain experts to extract the aspects of books.
\end{enumerate}
We have chosen the second method which can be evaluated at fine level of granularity, therefore we have asked the OpenEdition editorial team\footnote{http://www.openedition.org/}, which deals with  book reviews of social and human sciences, to enumerate the potential aspects that may be found in book reviews. They have listed the following aspects:

1. Book presentation

2. Problematic

3. Scientific context

4. Scientific method

5. Author's arguments

6. Book organization

7. judgment about the book

In each aspect one can find a various opinion targets which describe or name an aspect which can not be listed, the annotators will specify them during their annotation. 

\section{French Book Review Corpus Annotation}
For creating an annotated corpus of French book reviews, OpenEdition team and we have selected 200 book reviews in French language. We have automatically segmented  each review into sentences in order to annotate each sentence using Talismane\footnote{http://redac.univ-tlse2.fr/applications/talismane.html} syntax analyzer \cite{urieli_apport_2013}. The annotation should determine the 4 following elements: 
\begin{enumerate}
  \item 
Target: a word or an expression which one can express an opinion toward it.
\item 
Polarity: the expressed sentiment towards the target(positive, negative or neutral).
\item 
Polarity terms: the words which allow us to judge the expressed sentiment  (i.e. \textit{great} indicates positive sentiment).
\item 
Category: one of the previous seven categories identified by the editorial team. %In this work, this information will not be  exploited.
\item 
Occurrence: refers to the position of the target in the sentence. If the same target expression is repeated in the same sentence the first target occurrence is 1, the second repetition is 2 and so on.

\end{enumerate}
Three annotators have been asked to extract for each sentence all existing annotation elements. They have worked for 15 days, they annotated the same reviews, each one has annotated 7 reviews per day in average.
The following box shows a part of book review.\\
\fbox{    \parbox{12cm}{ {\scriptsize 
    Ce livre, version pour la publication d'un mémoire de DEA qui a reçu le prix Simone Genevois en 2002, est consacré à un sujet original et encore peu traité : le travail des conseillers historiques sur les films français des années 1970 et 1980. Une dizaine de films sont envisagés dans cette étude. Ce sont tous des films « historiques » français. L'ensemble reste malgré tout un peu hétéroclite puisque les deux films de René Allio considérés (les Camisards et Moi, Pierre Rivière…) ont été réalisés sans recours à ce genre de spécialiste, mais l'auteur s'en justifie par l'argument que les scénarios sont tirés d'ouvrages d'historiens renommés.  
   }
    }
}\\
The automatic segmentation divides this part into sentences, then the annotators extract the annotations for each sentence as the following box shows:\\
\fbox{
    \parbox{12cm}{  {\scriptsize 
{\color{red}{<review>}}

{\color{red}{<sentences>}}

<{\color{red}{sentence}} id="1">

<{\color{red}{text}}>

Ce livre , version pour la publication d' un mémoire de DEA qui a reçu le prix Simone Genevois en 2002 , est consacré à un sujet original et\_encore peu traité : le travail des conseillers historiques sur les films français des années 1970 et 1980 . 

<{\color{red}{/text}}>

<{\color{red}{Opinions}}>

<{\color{red}{Opinion}} {\color{blue}{target=}}" livre" {\color{blue}{category=}}"presentation" {\color{blue}{polarity=}}"positive" {\color{blue}{polarityterms=}}"original ; peu traite" {\color{blue}{occurrence=}}"1" />

<{\color{red}{/Opinions}}>

<{\color{red}{/sentence}}>

<{\color{red}{sentence}} id="2">

<{\color{red}{text}}>

Une dizaine de films sont envisagés dans cette étude . 

<{\color{red}{/text}}>

<{\color{red}{Opinions}}>

<{\color{red}{Opinion}} {\color{blue}{target=}}"films" {\color{blue}{category=}}"presentation" {\color{blue}{polarity=}}"neutre" {\color{blue}{polarityterms=}}"NULL" {\color{blue}{occurrence=}}"1" />

<{\color{red}{/Opinions}}>

<{\color{red}{/sentence}}>

<{\color{red}{sentence}} id="3">

<{\color{red}{text}}>

Ce sont tous des films " historiques " français .

<{\color{red}{/text}}>

<{\color{red}{Opinions}}>

<{\color{red}{Opinion}} {\color{blue}{target=}}"films" {\color{blue}{category=}}"presentation" {\color{blue}{polarity=}}"neutre" {\color{blue}{polarityterms=}}"NULL" {\color{blue}{occurrence=}}"1" />

<{\color{red}{/Opinions}}>

<{\color{red}{/sentence}}>

<{\color{red}{sentence}} id="4">

<{\color{red}{text}}>

L' ensemble reste malgré tout un\_peu hétéroclite puisque les deux films de René Allio considérés ( les Camisards et Moi , Pierre Rivière… ) ont été réalisés sans recours à ce genre de spécialiste , mais l' auteur s' en justifie par l' argument que les scénarios sont tirés d' ouvrages d' historiens renommés . 

<{\color{red}{/text}}>

<{\color{red}{Opinions}}>

<{\color{red}{Opinion}} {\color{blue}{target=}}"ensemble" {\color{blue}{category=}}"presentation" {\color{blue}{polarity=}}"negative" {\color{blue}{polarityterms=}}"heteroclite" {\color{blue}{occurrence=}}"1" />

<{\color{red}{Opinion}} {\color{blue}{target=}}"historiens" {\color{blue}{category=}}"methodology" {\color{blue}{polarity=}}"positive" {\color{blue}{polarityterms=}}"renommes" {\color{blue}{occurrence=}}"1" />

<{\color{red}{/Opinions}}>

<{\color{red}{/sentence}}>

<{\color{red}{/sentences}}>

<{\color{red}{/review}}>
    }
}
}\\

During the 15 days the three annotators have been annotated 97 common reviews. The first and second annotators have annotated 106 reviews while the third on has annotated 97.
Table 1 shows the statistics on the annotated book reviews. We firstly count the number of targets, categories, polarities given by each annotator which represent the first three lines in Table 1. Note that the number of targets is a bit different from the number of categories or polarities because  some sentences have been attributed to a category without determining a target and with or without the polarity. To measure the degree of agreement between the annotators, we have listed each possible combination among the three annotations, then the common targets, categories, polarities have been counted. We exclude 9 reviews when making the combination with the third annotator because he has annotated only 97 reviews.
\begin{table}[h!]
\begin{center}
\begin{tabular}{llccc}
\hline
 Annotater        &   Targets      &   Categories &   Polarities\\
\hline
 Annotater1      &3110& 3125& 3110\\
 
 Annotater2         &2976 &3028&2980\\
 
 Annotater3   & 3148 & 3164 &3132\\

  Annotater1+2 & 1294& 842& 1077\\
 Annotater1+3 & 1246 &717&981\\

 Annotater2+3 &1630 &965&1353\\
 Annotater1+2+3 &905 &410&647\\
\hline 
\end{tabular}
\caption{Statistics on book reviews annotations.}
\end{center}
\end{table}

From the last four lines of Table 1, we remark that the number of common targets and categories is very low comparing to those produced by each annotation. The reasons may be: the annotators have different viewpoints:
\begin{itemize}
   \item Some annotators extract a word or an expression as a target  others ignore it.
   \item Some annotators extract the same target but use different writing (e.g. "the man" vs "man").
 \end{itemize}    
The category is a bit confused for the annotators, they attribute different categories to the same text. Obviously, the common polarity number seems to be enough acceptable for the common targets.

\section{Opinion Target Extraction}
The objective of opinion target extraction is to extract all opinion target expressions in a book review, opinion target could be a word or multiple words. This extraction consists of the following steps:
\begin{enumerate}
  \item Review Segmentation
  
  This step segments each review into sentences.
  % In book reviews dataset, we have done that using Talismane syntax analyzer \cite{urieli_apport_2013}.
  \item Sentence Tokenizing
  
  Each sentence is tokenized to get the terms.
  % We can consider the spaces is the separators or use more complex tokenizer. We toknize each sentence using Talismane tokenizer which extracts the words, numbers and punctuations.
  \item Sentence Tagging
  
  Each term in the sentence should be tagged in order to be presented to a tagging classifier. We choose the IOB notation for representing each sentence in the review. Therefore, we distinguish the terms at the Beginning, the Inside and the Outside of opinion target.
For example, for the following review sentence:
\begin{center}
 \textit{"Mais la méthode avec laquelle il est présenté comme seule hypothèse recevable pose problème."}
 
\end{center} 
Where \textit{méthode} is a target. The tag of each word will be:
\begin{center}
 Mais:O  la:O  méthode:B  avec:O  laquelle:O  il:O est:O  présenté:O comme:O seule:O hypothèse:O recevable:O pose:O problème:O.
\end{center} 
  \item Feature Extraction
  
  This is the main step of opinion target extraction.  We extract the following features for each term in the sentence:
\begin{itemize}
  \item the term itself.

  \item term POS: We use  Talismane parser to attach a part of speech tag to each term.

 \item term shape: the shape of each character in the word (capital letter, small letter, digit, punctuation, other symbol)

 \item term type: the type of the word (uppercase, digit, symbol, combination )

 \item Prefixes (all prefixes having length between one to four ).

 \item Suffixes (all suffixes having length between one to four).

\end{itemize}
For each term in the sentence, we make use of three group of feature values:
\begin{enumerate}
  \item All the previous features for the term itself and the 2 and 3 previous and subsequent terms, respectively.
  \item the value of each two successive features in the the range -2,2 (the previous and subsequent two terms of actual word) for the following features:
word surface, word POS,word shape, word type.
\item We extract the value of each three successive features in the the range -1,1 for the two features: word POS and word.
\end{enumerate}
  \item Training Method
  
  We have used a Conditional Random Field (CRF) as we have done in opinion target extraction in restaurant reviews.
\end{enumerate}

\subsection{ Experiments and Results}
CRFsuite tool is used for this experiment with lbfgs algorithm. Table 2 shows the results of our experiments using different group of features.  The first line represents the experiment when using only the terms as  features which gives F1-score of 49.5\%. The second line word+POS makes use of the term and POS tagging as features which improves the results to reach 61.2\%.  The third line exploits the term, POS tags, types and shape features which improve the previous run by 0.3\%. The fourth line exploits all features including term, POS, shape, type and prefix and suffix which gives 61.5\%. Thus, it should note that the word and POS features seem to be enough to produce a good result. 
The last  line demonstrates the results obtaining from applying our system on the restaurant reviews provided by SemEval-2015 \cite{rosenthal_semeval-2015_2015}.
\begin{table}[h!]
\begin{center}
\begin{tabular}{|l|c|c|c|}
\hline
Experiment & Recall & Precision & F1-Score \\
\hline
word &0.8016& 0.4366 & 0.4958\\
word+pos &0.8272& 0.5375& 0.6102\\
w+pos+type+shape& 0.8015 &0.5434 &0.6131\\
w+pos+type+shape+pre+suf&0.8204& 0.5452& 0.6153\\
Restaurant& 0.5645 & 0.7268 &0.6355\\
\hline
\end{tabular}
\end{center}
\caption{The results of opinion target extraction in French book and English restaurant reviews.}
\end{table}

\subsection{Sentiment Polarity}
For a given set of opinion targets within a sentence, we should determine whether the polarity of each opinion target is positive, negative or neutral. For example, the system should extract the polarity of \texttt{méthode}  in the following sentence:
\begin{center}
  \textit{"Mais la méthode avec laquelle il est présenté comme seule hypothèse recevable pose problème."}
    méthode: negative
\end{center}
%This task can be seen as sentence level or phrase level sentiment Analysis. We have applied the same steps for sentiment polarity in restaurant reviews, before tokenizing we should firstly determine the context of opinion target, here we firstly detect the context, the context is the opinion target itself and all the surrounding terms enclosed between two separators like  (“,”, “;”, “!”), if another opinion target is also enclosed by these separators we consider it as a separator instead, and we do not take the terms after it or before it (according to its direction to the current opinion target). If the sentence has only an opinion target we take the whole sentence as context as explained for restaurant reviews.\\
%After the context detection, we should determine the polarity for each opinion target which could be positive, negative or neutral.
We propose to use a logistic regression  with the following features:
\begin{itemize}
  \item \textbf{Word n-grams Features}

Unigrams, bigrams and 3-grams are extracted for each word in the context without any stemming or stop-word removing, all terms with occurrence less than 3 are removed from the feature space.

\item \textbf{Z score Features}
 As described in  \cite{hamdan_impact_2014}, we tested different thresholds for choosing the words which have the highest Z score, a grid search in the interval [-2..5] with step of 0.5 has been done. We found -0.5 is the best one for book reviews. Thus, we added the number of words having Z score higher than -0.5 in each class positive,negative and neutral in the restaurant and laptop sets, the two classes which have the maximum number and minimum numbers of words having Z score higher than the threshold. These 5 features have been added to the feature space.

\end{itemize}

\subsection{Experiments}

We also trained a L1-regularized Logistic regression classifier implemented in LIBLINEAR. The classifier is trained on the training dataset using the previous features with the three polarities (positive, negative, and neutral) as labels.

We have used 10 fold cross-validation for evaluating our system. The results are shown in Table 3. The last two lines demonstrate the results obtaining from applying our system on the restaurant and laptop reviews provided by SemEval-2015 \cite{rosenthal_semeval-2015_2015}.

\begin{table}[h!]
\begin{center}
\begin{tabular}{|l|c|c|c|}
\hline
Experiment  & Accuracy Score \\
\hline
word& 70.87 \\
word+Z+3&70.50\\

word+Z+2& 71.66\\

word+Z+1.5& 72.31\\

word+Z+1&74.49\\

word+Z+0.5&77.94\\
word+Z+0.0&79.33\\
\textbf{word+Z-0.5}&\textbf{79.40}\\
word+Z-0.1&74.90\\
 Restaurant         & 75.5\\
 Laptop &  77.87\\
\hline
\end{tabular}
\end{center}
\caption{The results of sentiment polarity for opinion targets in French book, English restaurant and laptop reviews.}
\end{table}

The first line represents the experiment which exploits only the terms as features, it gives accuracy score  70\%. The remaining lines represent the experiments when exploiting the word and the Z score features, each line represents the same experiment but with a different Z threshold. We start by assigning 3 to Z score threshold and decrease this threshold until -1. The best result is given when using terms and Z score features with Z threshold of -0.5. The accuracy is 79\% which seems fair enough when comparing with the results produced in restaurant reviews (about 75.5\%).

\section{Related Work}
Aspect-Based Sentiment Analysis consists of several sub tasks. Some studies have proposed different methods for aspect detection and sentiment polarity analysis, others have proposed joint models in order to obtain the aspect and their polarities from the same model, these last models are generally unsupervised.

The early work on opinion target detection from on-line reviews presented by \cite{hu_mining_2004} used association rule mining based on Apriori algorithm \cite{agrawal_fast_1994} to extract frequent noun phrases as product features. For polarity detection, they used two seed sets of 30 positive and negative adjectives, then WordNet has been used to find and add the synonyms of the seed words. Infrequent product features or opinion targets had been processed by finding the noun related to an opinionated word.

Opinion Digger \cite{moghaddam_opinion_2010} also used Apriori algorithm to extract the frequent opinion targets. kNN algorithm is applied to estimate the aspect rating scaling from 1 to 5 stands for (Excellent, Good, Average, Poor, Terrible).

Supervised methods use normally  Conditional Random Fields (CRF) or Hidden Markov models (HMM). \cite{jin_novel_2009} applied a HMM model to extract opinion targets using the words and their part-of-speech tags in order to learn a model, then unsupervised algorithm for determining the opinion targets polarity using the nearest opinion word to the opinion target and taking into account the polarity reversal words (such as not).

A CRF model was used by \cite{jakob_extracting_2010} with the following features: tokens, POS tags, syntactic dependency (if the opinion target has a relation with the opinionated word), word distance (the distance between the word in the closest noun phrase and the opinionated word), and opinion sentences (each token in the sentence containing an opinionated expression is labeled by this feature), the input of this method is also the opinionated expressions, they use these expressions for predicting the opinion target polarity using the dependency parsing for retrieving the pair target-expression from the training set. We also applied a CRF model with different features \cite{hamdan_supervised_2014,hamdan_lsislif2015-1}.

Unsupervised methods based on LDA (Latent Dirichlet allocation) have been proposed. \cite{brody_unsupervised_2010} used LDA to figure out the opinion targets, determined the number of topics by applying a  clustering method, then they used a similar method proposed by \cite{hatzivassiloglou_predicting_1997} to extract the conjunctive adjectives, but not the disjunctive due to the specificity of the domain.

\cite{lin_weakly_2012}  proposed Joint model of Sentiment and Topic (JST) which extends the state-of-the-art topic model (LDA) by adding a sentiment layer, this model is fully unsupervised and it can detect sentiment and topic simultaneously.

\cite{wei_sentiment_2010} modeled the hierarchical relation between product aspects. They defined Sentiment Ontology Tree (SOT) to formulate the knowledge of hierarchical relationships among product attributes and tackled the problem of sentiment analysis as a hierarchical classification problem. Unsupervised hierarchical aspect Sentiment model (HASM) was proposed by  \cite{kim_hierarchical_2013} to discover a hierarchical structure of aspect-based sentiments from unlabeled online reviews.

\section{Conclusion and Future Work}
We have constructed a corpus of book reviews, segmented each review into sentences and asked three annotators to extract the opinion targets and their polarities in each sentence. We trained a CRF model for opinion target extraction and a logistic regression one for sentiment polarity. The obtaining results indicate that our systems perform as well as in restaurant reviews.

\bibliography{absabook}
\bibliographystyle{apalike}

\end{document}